\documentclass[10pt,journal,compsoc]{IEEEtran}
\usepackage[ruled,vlined]{algorithm2e}
\usepackage{url}
\usepackage{array}
\usepackage{tabularx}

\usepackage{graphicx}
\usepackage{color}
\usepackage{soul}
\usepackage[dvipsnames]{xcolor}
\usepackage{amssymb}
\usepackage{amsmath}
\ifCLASSOPTIONcompsoc
  \usepackage[nocompress]{cite}
\else
  \usepackage{cite}
\fi
\ifCLASSINFOpdf
\else
\fi
\begin{document}

\title{MultiSAGE: a multiplex embedding algorithm \\for inter-layer link prediction}

\author{Luca~Gallo,
        Vito~Latora,
        and~Alfredo~Pulvirenti%
\IEEEcompsocitemizethanks{\IEEEcompsocthanksitem L. Gallo and V. Latora are with the Department of Physics and Astronomy, University of Catania, 95125 Catania, Italy, and INFN Sezione di Catania, Via S. Sofia, 64, 95125 Catania, Italy\protect\\
E-mail: luca.gallo@phd.unict.it
\IEEEcompsocthanksitem L. Gallo is with naXys, Namur Institute for Complex Systems, University of Namur, 5000 Namur, Belgium.
\IEEEcompsocthanksitem V. Latora is with School of Mathematical Sciences, Queen Mary University of London, London E1 4NS, UK.
\IEEEcompsocthanksitem A. Pulvirenti is with Department of Clinical and Experimental Medicine, University of Catania, 95125 Catania, Italy.}%
\thanks{Manuscript received ; revised .}}

\IEEEtitleabstractindextext{%
\begin{abstract}
Research on graph representation learning has received great attention in recent years. However, most of the studies so far have focused on the embedding of single-layer graphs. The few studies dealing with the problem of representation learning of multilayer structures rely on the strong hypothesis that the inter-layer links are known, and this limits the range of possible applications. Here we propose MultiSAGE, a generalization of the GraphSAGE algorithm that allows to embed multiplex networks. We show that MultiSAGE is capable to reconstruct both the intra-layer and the inter-layer connectivity, outperforming GraphSAGE, which has been designed for simple graphs. Next, through a comprehensive experimental analysis, we shed light also on the performance of the embedding, both in simple and in multiplex networks, showing that either the density of the graph or the randomness of the links strongly influences the quality of the embedding.
\end{abstract}

}

\maketitle

\IEEEdisplaynontitleabstractindextext
\IEEEpeerreviewmaketitle

\IEEEraisesectionheading{\section{Introduction}\label{sec:introduction}}
\IEEEPARstart{G}{raphs} (or networks) are mathematical data structure to represent real-world complex systems. In the most general view, a graph consists of a  entities or objects (i.e., nodes), together with a set of relations (i.e., edges) between pairs of such entities. Examples of graphs include: social networks~\cite{scott1988social} in which entities are people and links represent friendship; co-authorship and citation networks~\cite{liu2005co}; biological networks~\cite{barabasi2011network} where nodes are for example proteins and relations between nodes may represent functional or physical interactions; knowledge graphs~\cite{ji2021survey} (also known as semantic networks) in which nodes are real-world entities, such as objects, events, situations and edges are relationship between them. 
In many contexts, simple graphs, where there is at most one edge between each pair of nodes are enough to model the application.
However, graphs having different types of relations allow to model, in a more comprehensive way, the complex system under study.
Such kind of graphs are called multi-relational which are commonly  divided in two important sub-groups known as heterogeneous and multiplex graphs. In heterogeneous graphs, nodes have types, and therefore, we can partition the set of nodes into disjoint sets. Whereas, multiplex graphs can be decomposed in a set of $k$ layers. Every node belongs to every layer (each layer allows to represents a specific kind of relation), along with intra-layer edges type for each layer. Furthermore, inter-layer edges types can exist, and these connect the same node across different layers. 

Due the ubiquity of networks in the real world, graph analysis has become mandatory in several domains \cite{zhou2020graph}. In particular, the field of graph representation has received an impressive research interest the past few years \cite{bacciu2020gentle,chen2020graph,grohe2020word2vec}. Network representation learning or low dimensional network embedding, consist of algorithmic methodologies, which allow to project the nodes of a network into a multidimensional space preserving the structure of the network and its intrinsic properties. 
The usage of dimensionality reduction techniques to encode, into vectors, the high-dimensional information of nodes' graph neighborhood allows then to apply traditional Machine Learning methodologies to networks. Indeed, such node embeddings can then be used into downstream learning tasks and analysis such as: node classification~\cite{rong2019dropedge}, link prediction~\cite{lu2011link}, and community detection~\cite{fortunato2010community}. 

The performance of machine learning methods crucially depend on the quality of the vector representations. Therefore, there is a wealth of research proposing a wide range of vector-embedding methods for various applications. Seminal works in this area include LINE~\cite{tang2015line}, DeepWalk~\cite{perozzi2014deepwalk} and node2vec~\cite{grover2016node2vec} which are based on taking short random walks in a graph and interpreting the sequence of nodes seen on such random walks as if they were words appearing together in a sentence. However, from a deep learning perspective, such  embedding methods are all shallow, indeed they directly optimise the output vectors without hidden layers.
Although, shallow approach can be generalized to multi-relational graphs it presents several limitations. Shallow embedding methods are transductive. Indeed, these methods are only able to generate embeddings for nodes that were present during the training phase. 

To go beyond such limitations, shallow encoders can be replaced with more sophisticated encoders that take into account the structure and attributes of the graph. Among the most popular encoders we have graph neural networks (GNNs). The key feature of a GNN is that it uses a kind of neural message passing, which essentially is a generalization of the Weisferler-Leman test \cite{shervashidze2011weisfeiler,leman1968reduction}. At each iteration, every node aggregates information from its local neighborhood. As these iterations go on, each node embedding contains more information obtained from remote nodes of the graph.
In the last few years, research in this area has been proposed in this context \cite{grohe2020word2vec,chen2020graph,velivckovic2017graph,liu2020towards,9432709}. One of the most prominent example of an inductive node-embedding tool based on GNNs is GraphSage \cite{hamilton2017inductive}.

More recently, research models for multi-layer networks embedding have raised.
In \cite{cen2019representation} authors propose an embedding approach  for multigraphs, in \cite{gong2020heuristic} a generalization of  deep-walk and node2vec for multiplex graphs were introduced. \cite{gong2020heuristic}. Other relevant approaches include \cite{liu2017principled,qu2017attention,zhang2018scalable,shi2018mvn2vec}. However, in all of these approach no distinction between inter- and intra-layer connections is taken into account.

In this paper, we introduce MultiSAGE, a generalization of the GraphSAGE algorithm for embedding multiplex networks. Our embedding approach focuses only on unlabeled graphs but can be easily generalized to labeled graphs. Through a comprehensive experimental analysis based on three benchmark datasets, we shed light also on the limit of the vector embedding in producing reliable vector embedding, both for simple graphs and multiplex. Indeed, our analysis shows that either density of the graphs or the randomness of the links actually limit the quality of the embedding.

\section{The MultiSAGE algorithm}

\subsection{Multiplex networks}
A multiplex network is a generalization of a network that allows to naturally represent systems of connected units when multiple types of interactions among them exist \cite{de2013mathematical,battiston2014structural}. Multiplex networks  can be see as a particular type of a multilayer networks \cite{kivela2014multilayer,boccaletti2014structure}. We will first define them describing their graph and matrix representation, and presenting the crucial concept of \emph{intra-layer} and \emph{inter-layer} links, as well as the one of \emph{supra-adjacency matrix} \cite{cozzo2018multiplex}. 

\begin{figure}[t]
    \centering
    \includegraphics[width=\linewidth]{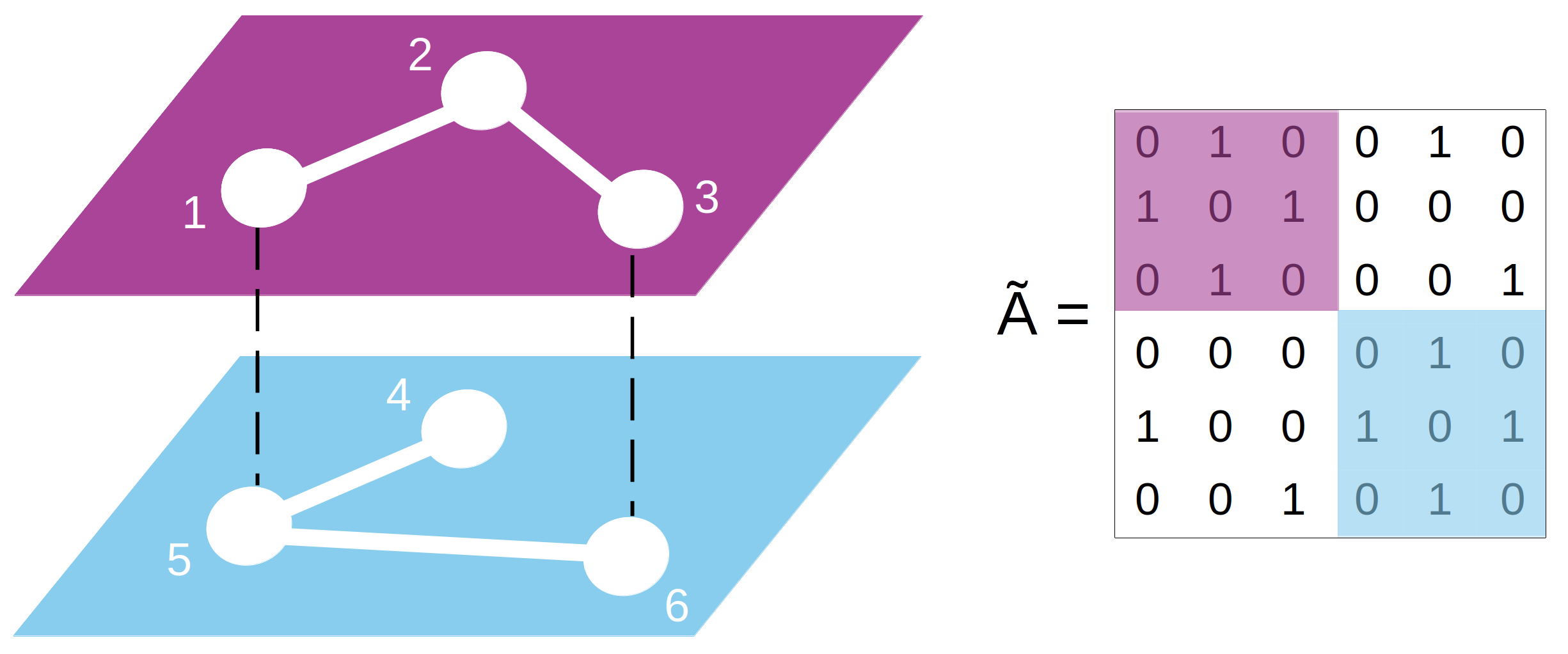}
    \caption{A multiplex network with two layers (left) and its supra-adjacency matrix $\tilde{\rm A}$ (right). Intra-layer links are colored in white, and correspond respectively to the purple and blue diagonal blocks of $\tilde{\rm A}$. Inter-layer links are displayed as black dashed lines, and correspond to white off-diagonal blocks of $\tilde{\rm A}$.}
    \label{fig:multiplex_scheme}
\end{figure}

Let us consider a set $\mathcal{V}$ of $N$ nodes interacting through $L$ different types of relations, represented as $L$ different layers. 
Let us assume that $N_\alpha$ nodes lie on each layer $\alpha$, with $\alpha=1,\dots, L$ so that $N_1+N_2+\dots+N_L=N$. Each layer of the multiplex consists in a network $G_\alpha = (\mathcal{V}_\alpha,\mathcal{E}_\alpha)$, where $\mathcal{V}_\alpha \subseteq \mathcal{V}$ is the set of $N_\alpha$ nodes in the layer, while $\mathcal{E}_\alpha$ is a set of edges representing the relations of type $\alpha$. Hereby, we will refer to the edges in $\mathcal{E}_{\mathrm{intra}} = \bigcup_{\alpha=1}^{L} \mathcal{E}_\alpha$ as \emph{intra-layer links}. The connectivity of each layer $G_\alpha$ can be encoded as an adjacency matrix $A^\alpha\in\mathbb{R}^{N_\alpha\times N_\alpha}$, whose element $a_{ij}^\alpha$ is equal to one, if $(i,j)\in \mathcal{E}_\alpha$, i.e., nodes $i$ and $j$ interacts through the relation $\alpha$, while it is equal to zero otherwise.

In addition to the intra-layer connectivity, we assume to connect some of the nodes lying on different layers, so that given $\alpha,\beta$, with $\alpha\neq\beta$, a node lying on layer $\alpha$ can be linked with at most one node on layer $\beta$. This constraint makes the multiplex a suitable representation for those systems where the same set of individuals are connected through different relations \cite{boccaletti2014structure}. In what follows, we will call these connections \emph{inter-layer links}. As a further hypothesis, we assume that if a node $i$ on layer $\alpha$ is connected to a node $j$ on layer $\beta$, and if $j$ is connected to a node $k$ on layer $\gamma$, then nodes $i$ and $k$ are also connected. Therefore, the graph $G_\mathrm{inter} = (\mathcal{V},\mathcal{E}_\mathrm{inter})$, where $\mathcal{E}_\mathrm{inter}$ is the set of inter-layer links, will be formed by disconnected components that are either cliques or isolated nodes \cite{cozzo2018multiplex}. Such a graph can be as well characterized by an adjacency matrix $C\in\mathbb{R}^{N\times N}$, whose element $c_{ij}$ is equal to one, if $(i,j)\in \mathcal{E}_\mathrm{inter}$, whereas it is equal to zero otherwise.

Finally, we can define the \emph{supra-adjacency matrix} $\Tilde{A}$, which encodes both intra-layer and inter-layer connections. By labeling the nodes according to the labels of the layer, i.e., indices from $1$ to $N_1$ denotes the nodes in the first layer, from $N_1+1$ to $N_1+N_2$ the nodes in the second layer, and so on, we can write the supra-adjacency matrix as
\begin{equation}
    \Tilde{\rm A} = \bigoplus\limits_{\alpha} A^\alpha + C,
\end{equation}
where $\bigoplus$ denotes the direct sum. An example of a multiplex network and its supra-adjacency matrix is shown in Fig.~\ref{fig:multiplex_scheme}. Note that the diagonal blocks correspond to the adjacency matrices of the graphs at each layer, thus encoding the intra-layer connectivity, while the off diagonal blocks contain the information on the inter-layer connections. It is worth mentioning that we can interpret $\Tilde{\rm A}$ as the adjacency matrix of a single-layer multigraph $\mathcal{M} = (\mathcal{V}, \mathcal{E})$, where $\mathcal{E}=\mathcal{E}_{\rm intra} \cup \mathcal{E}_{\rm inter}$, composed of $N$ nodes where there are two kinds of links, namely the intra-layer and the inter-layer. We will call such a graph the \emph{flattened multiplex network}. This will come up particularly useful when comparing our algorithm to GraphSAGE.

Let us remark that our definition is more flexible compared to the one usually given for multiplex networks. Indeed, it is common to assume that each layer of the multiplex is composed by the same number of nodes, i.e., $N_\alpha=m,\,\forall \alpha$, and that for any $\alpha,\beta$, a node on layer $\alpha$ is connected to \emph{exactly} one node on layer $\beta$ \cite{boccaletti2014structure, battiston2014structural}. Here, to extend the range of possible applications of our algorithm, we do not consider such extra constraints. As an example, let us consider the case of different online social networks (OSNs), which can be represented by a single entity, namely a multiplex network where each layer encodes the connections in one of the OSNs. If a user can have only one account for each OSN, we constraint ourselves to the case of a multiplex. In such an example, assuming that each layer is composed by the same number of nodes and that all the possible inter-layer links exist is too restrictive. First, the users might not have an account on every OSN. Second, these accounts might not be connected one another, i.e., it is not know \emph{a priori} that two accounts on different OSN are owned by the same user. Hence, relaxing the constraint of having all the possible inter-layer connections opens to relevant applications like, for instance, the one of de-anonymization \cite{ji2016graph}, which can be seen as an inter-layer link prediction.  

\subsection{The algorithm}
The embedding algorithm MultiSAGE represents the natural generalization of the GraphSAGE algorithm \cite{hamilton2017inductive} to multiplex networks. In order to provide a vector representation $\mathbf{z}_n$ for a graph node $n$, GraphSAGE relies on the idea of training a set of $K$ \emph{aggregator functions} $f_k$, with $k\in \{1,\ldots,K\}$, that learn to aggregate information from the node $K$-th neighborhood. In other words, given a node $n$, its embedding vector depends, according to a first aggregator function, on the features of its first-neighbors, which in turns are evaluated, through a second aggregator function, from the characteristics of their first-neighborhoods, i.e. the second-neighborhood of $n$, and so on, up to a certain depth $K$. Therefore, the vector $\mathbf{z}_n$ can be determined once all the aggregator functions are trained. 

In the original version of the GraphSAGE algorithm, which deals with the embedding of (single-layer) graphs, when considering a specific node $n$, each aggregator $f_k$ is a function of the features of the node itself and of the ones of its first-neighborhood. Conversely, since for multiplex networks we can distinguish, for each node, both an intra-layer and an inter-layer neighborhood, in the MultiSAGE embedding algorithm the aggregator functions will depend on the features of node $n$, and of its intra-neighbors, but also on the features of its inter-neighbors. The MultiSAGE embedding generation is formally described by Algorithm \ref{alg:multisage}.

\begin{algorithm}[t]
\SetKwInOut{Input}{Input}\SetKwInOut{Output}{Output}
\Input{Flattened multiplex network $\mathcal{M}=(\mathcal{V},\mathcal{E})$; input feature vector $\mathbf{x}_n$, $\forall n \in \{1,\dots,|\mathcal{V}|\}$; depth $K$; 
differentiable aggregator functions $f_k$, $\forall k\in\{1,\ldots,K\}$;
intra-layer and inter-layer neighborhood functions ${N}_{H},{N}_{V}: \mathcal{V} \rightarrow \mathcal{P}(\mathcal{V})$} 
\Output{Vector representation $\mathbf{z}_n$, $\forall n \in \{1,\dots,|\mathcal{V}|\}$}
\BlankLine
$\mathbf{h}_{n}^{0}\leftarrow \mathbf{x}_n$, $\forall n \in \{1,\dots,|\mathcal{V}|\}$\;
\For{$k=1,\dots,K$}{\label{for_deep}
\For{$n=1,\dots,|\mathcal{V}|$}{
$\mathbf{h}_{n}^{k}\leftarrow f_k(\{\mathbf{h}_{m}^{k-1}, \forall m \in \mathcal{N}_H(n)\},\{\mathbf{h}_{m}^{k-1}, \forall m \in \mathcal{N}_V(n)\},\mathbf{h}_{n}^{k-1})$\;
}
}
$\mathbf{z}_{n}\leftarrow \mathbf{h}^{K}_{n}$, $\forall n \in \{1,\dots,|\mathcal{V}|\}$\;
\caption{MultiSAGE embedding generation algorithm}\label{alg:multisage}
\end{algorithm}

Again, the core idea behind the embedding algorithm is to generate, for each node of the multiplex network, a vector representation by aggregating the features of the node $K$-th neighborhood, through a series of $K$ iterations. This time, however, at each iteration $k\in\{1,\ldots,K\}$, each node $n$ is represented by a feature vector $\mathbf{h}_n^{k}\in\mathbb{R}^{d_k}$, which is evaluated by aggregating, through the function $f_k$ the representation vectors of the intra-neighbors of node $n$, $\{\mathbf{h}_{m}^{k-1}, \forall m \in \mathcal{N}_H(n)\}$, of the inter-neighbors of $n$, $\{\mathbf{h}_{m}^{k-1}, \forall m \in \mathcal{N}_V(n)\}$, and of the node $n$ itself, $\mathbf{h}_{n}^{k-1}$, at the previous step, i.e. $k-1$. Note that the subscripts $H$ and $V$ stand for \emph{horizontal} and \emph{vertical} respectively. The algorithm is initialized by defining the representation vector $\mathbf{h}_n^0$ at step $k=0$ as the input node features $\mathbf{x}_n$. Finally, the embedding vector $\mathbf{z}_n$ of node $n$ is given by the representation vector at step $K$, i.e. $\mathbf{h}_n^K$. Note that, in general, the dimension $d_k$ of the representation vectors can be different from one step to another. Lastly, we remark that, differently from GraphSAGE, in Algorithm \ref{alg:multisage}  we do not consider a concatenation step, so that the vector $\mathbf{h}_{n}^{k-1}$ contributes to the aggregation step through the function $f_k$.

The aggregation of the representation vectors can be performed employing a variety of aggregator functions $f_k$. In this paper, we focus on the simplest version of GraphSAGE, namely the one using a weighted mean to define the aggregator functions. In this case, we can substitute the line of pseudo-code relative to $f_k$ in Algorithm \ref{alg:multisage} with
\begin{equation}
\begin{aligned}
     \mathbf{h}_{n}^{k}\leftarrow \theta\Bigg(& \mathcal{W}_{H}^k\sum\limits_{m\in\mathcal{N}_{H}(n)}\frac{\mathbf{h}_{m}^{k-1}}{|\mathcal{N}_{H}(n)|}\\
     & +\mathcal{W}_{V}^k\sum\limits_{m\in\mathcal{N}_{V}(n)}\frac{\mathbf{h}_{m}^{k-1}}{|\mathcal{N}_{V}(n)|}+\mathcal{S}^k \mathbf{h}_{n}^{k-1}\Bigg),
     \label{eq:mean_aggregator}
     \end{aligned}
\end{equation}
where $\mathcal{W}_{H}^k, \mathcal{W}_{V}^k, \mathcal{S}^k \in \mathbb{R}^{d_{k},d_{k-1}}$ are matrices weighting the contribution of the intra-neighborhood, of the inter-neighborhood, and of the focal node, respectively, while $\theta$ is a nonlinear activation function. The representation vector $\mathbf{h}_n^{k}$ of node $n$ at step $k$ is determined by three factors, namely the mean of the representation vectors at step $k-1$ of its intra-neighbors, the mean of the representation vectors at step $k-1$ of its inter-neighbors, and the representation vector at step $k-1$ of the node itself, all of which are multiplied by the respective weight matrices. Given this choice for the aggregator functions, once the algorithm is provided with the activation function $\sigma$, the only unknown variables that have to be determined through the training process are the weight matrices $\mathcal{W}_{H}^k$, $\mathcal{W}_{V}^k$ and $\mathcal{S}^k$.

Consistently with the GraphSAGE algorithm, to learn the weight matrices defining the MultiSAGE aggregator functions in a unsupervised setting, we introduce a loss function based on the supra-adjacency matrix of the multiplex network. In particular, the loss function $J_{\mathcal{M}}$ is defined so to force neighboring nodes (both intra- and inter-layer) to have similar embedding vectors, while constraining nodes that are not connected, i.e. the negative links, to have dissimilar representations. However, since the number of negative links can be high, the computation of the loss function for large networks can be computationally expensive. To reduce the computational cost of the learning process, it is common to rely on \emph{negative sampling} \cite{mikolov2013distributed,yang2020understanding}, namely to evaluate the loss function only on a random subset of all the possible negative links. Formally, given the embedding vectors $\mathbf{z}_n$, $\forall n\in V$, we define
\begin{equation}
\begin{aligned}
    J_{\mathcal{M}} = - \sum\limits_{(n,m)\in \mathcal{E}} \Big\{ & \mathrm{log}(\sigma(\mathbf{z}_n^\top\mathbf{z}_m)) \\
    & + \sum\limits_{\overline{m}\sim P(n)}\mathrm{log}(\sigma(-\mathbf{z}_n^\top\mathbf{z}_{\overline{m}}))\Big\},
    \end{aligned}
\end{equation}
where $P(n)$ is the negative sampling probability distribution, and $\sigma(x)$ is the sigmoid function $\sigma(x) = 1/(1+e^{-x})$. Note that the sum in the curl brackets has $Q$ terms, corresponding to the number of negative samples.

\section{Data and evaluation}
\subsection{Multiplex network data}

\begin{table}[]
    \centering
\begin{tabular}{c|c|c|c|c}
Network & Nodes & Layers & Intra-layer links & Inter-layer links \\
\hline
arXiv & 19310 & 13  & 48657 & 20738 \\
Drosophila & 11867 & 7 & 40228 & 5173 \\
ff-tw-yt & 11827 & 3 & 74815 & 6028 \\
\end{tabular}
\label{tab:datasets}
\caption{Basic features of the multiplex networks emplyed to test the performance of the MultiSAGE algorithm.}
\end{table}

\begin{table*}[t!]
    \centering
\begin{tabular}{p{2.5cm}|c@{\hspace{1em}}c|c@{\hspace{1em}}c|c@{\hspace{1em}}c}
Algorithm & \multicolumn{2}{c|}{ff-tw-yt} & \multicolumn{2}{c|}{Drosophila} & \multicolumn{2}{c}{arXiv} \\
& intra & inter & intra & inter & intra & inter\\ 
\hline
GraphSAGE & $0.47 \pm 0.02$ & $0.56 \pm 0.02$ & $0.54 \pm 0.01 $ & $0.63 \pm 0.02$ & $0.72 \pm 0.01$ & $0.70 \pm 0.01$ \\
&\multicolumn{2}{c|}{}&\multicolumn{2}{c|}{}&\multicolumn{2}{c}{}\\
MultiSAGE & $0.48 \pm 0.02$ & $0.62 \pm 0.02$ & $0.51 \pm 0.02$ & $0.77 \pm 0.02 $ & $0.70 \pm 0.02$ & $0.83 \pm 0.01$ \\
\end{tabular}
\caption{Intra-layer and inter-layer link prediction results for both MultiSAGE and GraphSAGE on the three datasets. The table reports the average AUC, along with the standard deviation.}
\label{tab:results_complete}
\end{table*}

To test the performance of the MultiSAGE algorithm in predicting both intra-layer and inter-layer connections, we have analyzed different types of real-world multiplex networks, such as collaboration, biological, and online social networks, as well as synthetic datasets. For each empirical dataset, we consider only the largest connected component, and we ignore, if these are given, the directions and the weights of the links, i.e., we assume all the networks to be undirected and unweighted.

\textbf{arXiv} \cite{de2015identifying}. The arXiv multiplex network is a collaboration network where each layer represents a different category, i.e. research topic, of the pre-print archive. To generate the network, only the papers including the word ``networks'' in the title or in the abstract published before May 2014 were selected. The network largest connected component consists in this case of $19310$ nodes over $13$ layers, connected through $48657$ intra-layer connections, and $20738$ inter-layer ones. 

\textbf{Drosophila} \cite{de2015structural,stark2006biogrid}. This is the protein-genetic interaction multiplex network of the common fruit fly \emph{Drosophila melanogaster}, where layers correspond to interactions of different nature. The dataset is gathered from the Biological General Repository for Interaction Datasets (BioGRID), updated to January 2014. The largest connected component consists of $118670$ nodes over $7$ layers, with $40228$ intra-layer links, and $5173$ inter-layer ones. 

\textbf{ff-tw-yt} \cite{DickisonMagnaniRossi2016}. This is a multiplex network obtained from Friendfeed (ff), a social media aggregator, which allows the users to register their accounts on other online social networks (OSNs). The retrieved multiplex network consists of users who registered a single Twitter (tw) account and a single YouTube (yt) account on Friendfeed, and whose Twitter and YouTube accounts are associated to one Friendfeed account. The largest connected component consists of $11827$ nodes over $3$ layers, with $74815$ intra-layer connections, and $6028$ inter-layer ones.

The main characteristics of the datasets considered in this study are reported in Table \ref{tab:datasets}. It is worth remarking that, for all the multiplex networks described above, no external feature vectors is provided for the nodes. Therefore, for each node $n$ we will consider as the input feature vector $\mathbf{x}_n$ a one-hot encoder vector \cite{kipf2016semi}, i.e., $x_{n,i} = \delta_{ni}$, where $\delta$ is the Kronecker delta.

\subsection{Experimental setup and evaluation}

The main task we want to test MultiSAGE on is that of predicting both the intra-layer and the inter-layer links of a multiplex network. To assess the performance of our algorithm, we consider the following experimental setup. We first select a random sample of $20\%$ of the network nodes, which we refer to as \emph{marked nodes}. We then define the \emph{test} and the \emph{training} sets. Each one of these sets consists of positive examples, i.e., the links of the network, and negative examples, i.e., couples of unconnected nodes. As positive examples in the test set, we consider a subset of $20\%$ of the intra-layer links and all the inter-layer links of the marked nodes. Positive examples of the training set consist of all the remaining intra- and inter-layer links of the entire network. As concerns the negative examples, we include in the test set $20\%$ of all the possible negative intra-layer links among the marked nodes, and all the possible inter-layer links among them. The remaining uncoupled pair of nodes in the entire network form the negative examples of the training set.

Since the algorithm is trained so that adjacent nodes have a similar vector representation, while nonadjacent nodes have a dissimilar one, we would expect MultiSAGE to discriminate between positive and negative links in terms of the similarity of their vertices. Indeed, when it performs well, the similarity between the vertices of a positive link should be higher than the similarity between the vertices of a negative link. Instead of setting an arbitrary discrimination threshold on the value of the vertex similarity, so that all links whose vertex similarity is above threshold are considered true links of the networks and vice versa, to estimate the algorithm performance we rely on the Receiving Operating Characteristic (ROC) curves. To construct them, for both intra-layer and inter-layer links, we evaluate the frequency distribution of the vertices similarity for positive and negative links respectively, and estimate how the true positive rate, namely the fraction of positive links correctly predicted as true links, and the false positive rate, i.e., the fraction of negative links incorrectly predicted, vary as a function of the discrimination threshold. Therefore, as a scalar measure of the algorithm performance, we consider the area under the ROC curve (AUC), which corresponds to the probability that a randomly chosen positive link have a vertex similarity higher than one of a randomly chosen negative link. Therefore, when the algorithm is able to distinguish between positive and negative links, the value of the AUC tends to $1$, while in the opposite case it goes to $0.5$, i.e., random classifier. 

\section{Results}

\begin{figure*}[t]
    \centering
    \includegraphics[width=0.49\linewidth]{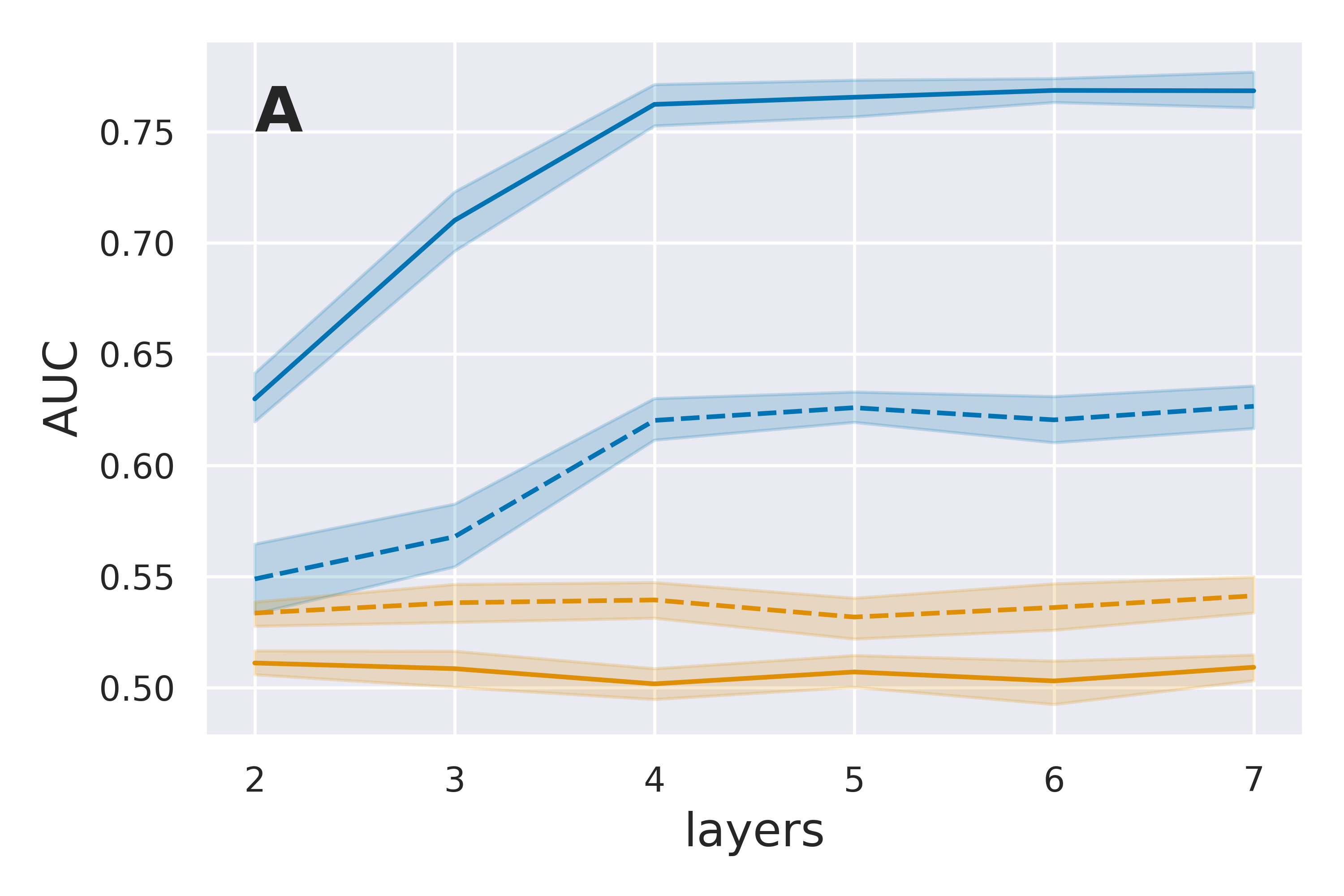}
    \includegraphics[width=0.49\linewidth]{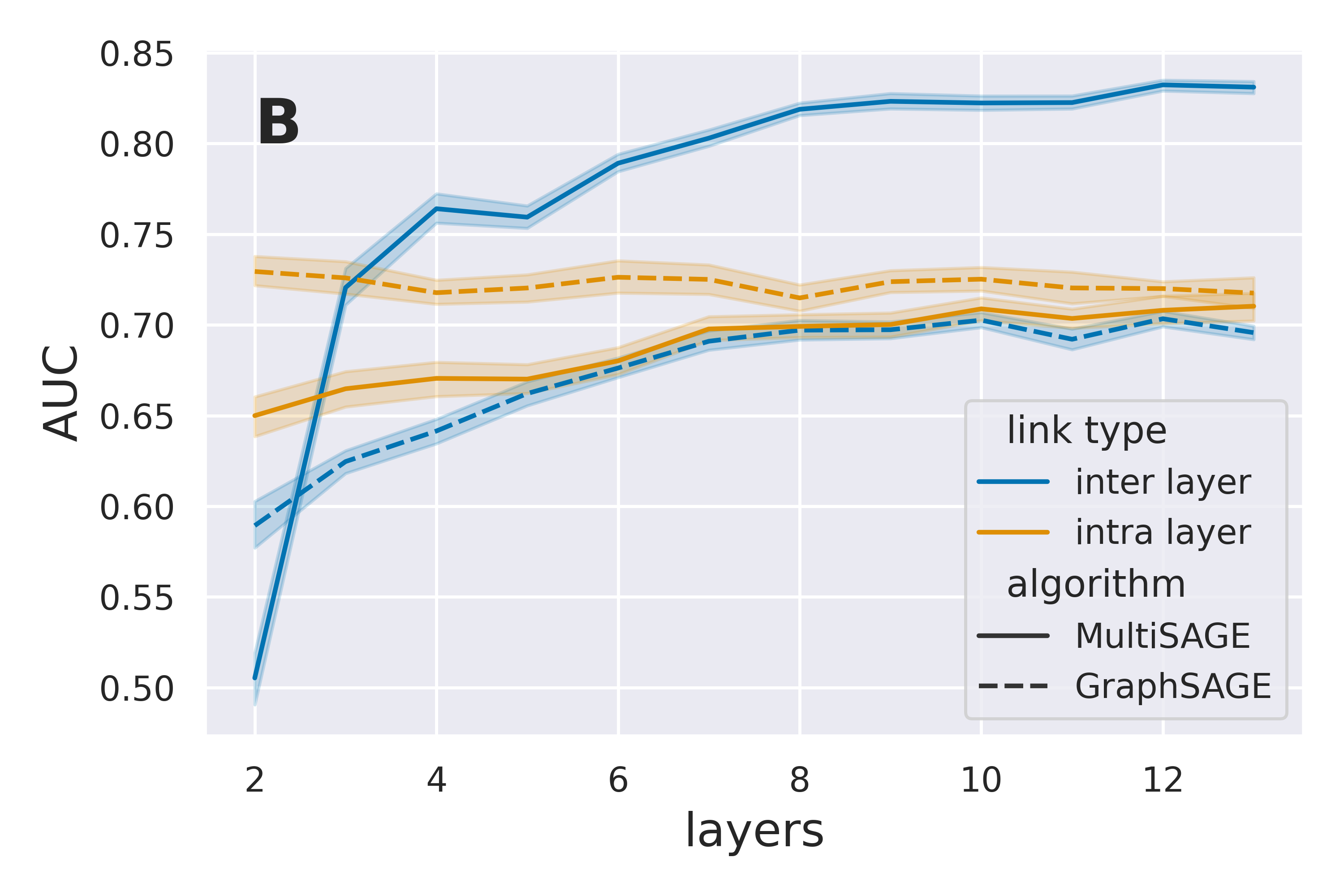}
    \caption{Variation of the AUC for inter-layer (blue lines) and the intra-layer (orange lines) link prediction as a function of the number of layers, for MultiSAGE (solid lines) and GraphSAGE (dashed lines), on A) Drosophila and B) arXiv multiplex networks.}
    \label{fig:lasagna_results}
\end{figure*}

In this section we present a series of results obtained for GraphSAGE and our generalization to multiplex network. First, we compare the performance of the two algorithms in predicting the intra-layer and inter-layer connectivity of the multiplex networks described above. Then, we analyze the dependence of both MultiSAGE and GraphSAGE on the number of layers of a network, assessing how the performances vary as we gradually change the number of layers to embed. Finally, we will further investigate such a dependence, showing how the prediction accuracy of GraphSAGE is related to the link density and to the randomness of the network. 

\subsection{Embedding multiplex networks}
Our first analysis consists in evaluating the performance of MultiSAGE in predicting intra-layer and inter-layer links when embedding a multiplex network. To do so, for each of the three dataset illustrated, we consider the experimental setup described in the previous section, training and testing the algorithm on $20$ different samplings of the marked nodes and of the train and test sets. As a measure of the algorithm performance, we consider the AUC averaged over the different realizations of the embedding procedure. We use the standard deviation as a measure of the error. As a benchmark, we compare the results obtained by MultiSAGE with the one of GraphSAGE, where we have trained the algorithm with no distinction between intra-layer and inter-layer links. 

Table \ref{tab:results_complete} illustrates the results obtained by the two algorithms, on the three datasets, for both intra-layer and inter-layer link prediction. First, we note that, for both GraphSAGE and MultiSAGE, the value of the average AUC tends to be generally higher for the inter-layer link prediction than for the intra-layer link prediction. On all the multiplex networks considered, MultiSAGE provides significantly better predictions of the inter-layer connectivity compared to GraphSAGE, while the two algorithms have the same performance, within the error bars, when reconstructing the intra-layer connections. Furthermore, we remark that the performance of both algorithms seem to positively depend on the number of layers of the network. Indeed, when predicting the intra-layer links, both MultiSAGE and GraphSAGE obtain results that are comparable to those of a random classifier, i.e., $\mathrm{AUC}=0.5$, on the ff-tw-yt network, which has three layers, and on the Drosophila network, composed by seven layers. Conversely, on the arXiv network, which is formed by thirteen layers, the value of the AUC goes up to abouth $0.70$ for both algorithms. As regards the inter-layer link prediction, the dependency on the number of layers seems to be more pronounced, with the value of AUC ranging from $0.62$ and $0.56$ for the online social network, to $0.83$ and $0.70$ for the collaboration network, for MultiSAGE and GraphSAGE respectively.

Overall, for both intra-layer and inter-layer link prediction, the number of layers of the network seems to have a positive impact on the performance of the algorithms, so that the more the layers the higher the value of the AUC. In the next section, we will provide an in-depth analysis of the dependence of the algorithm performance on the number of layers of the network, introducing an explanatory measure for the accuracy of the link prediction.

\begin{figure*}[t]
    \centering
    \includegraphics[width=0.49\linewidth]{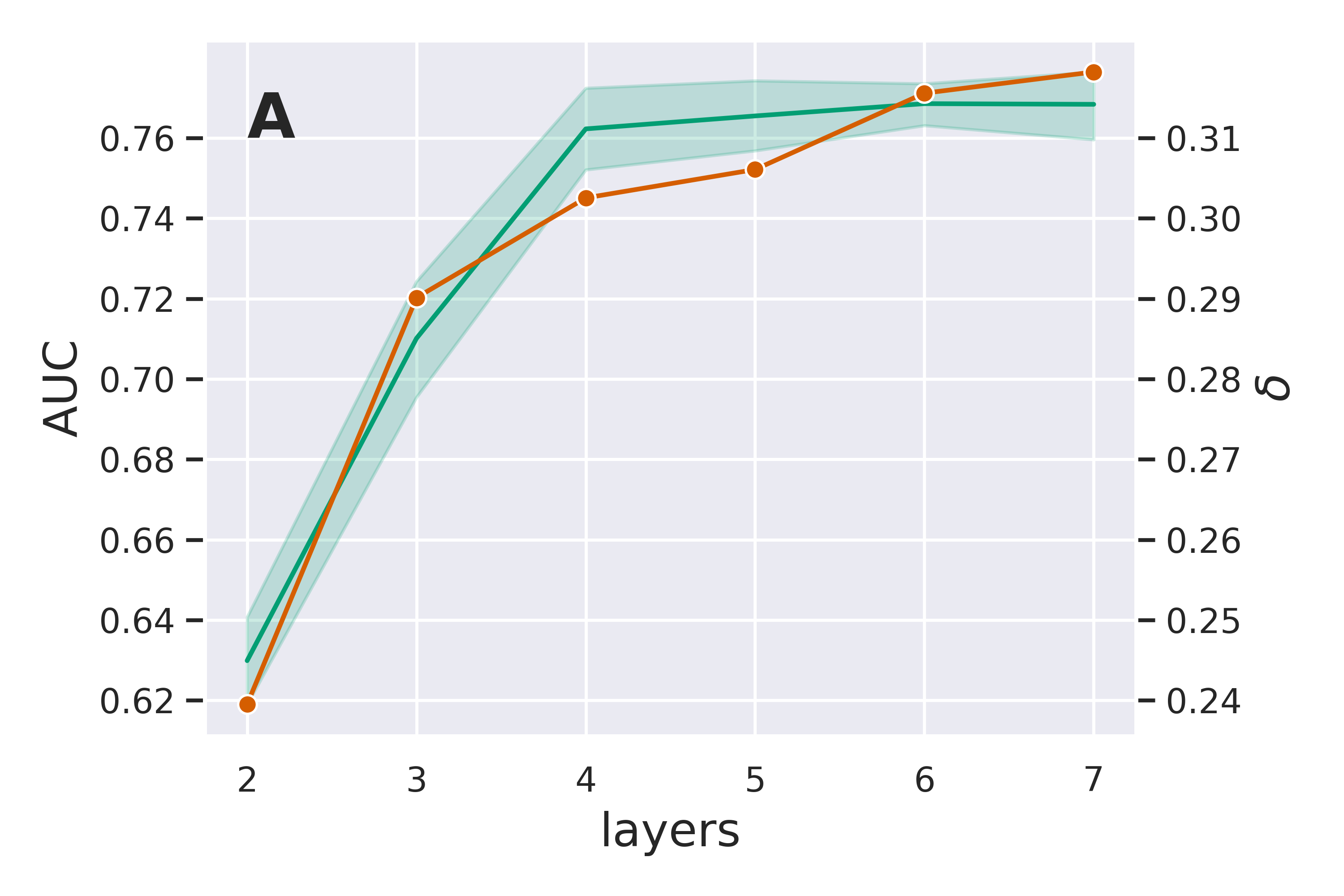}
    \includegraphics[width=0.49\linewidth]{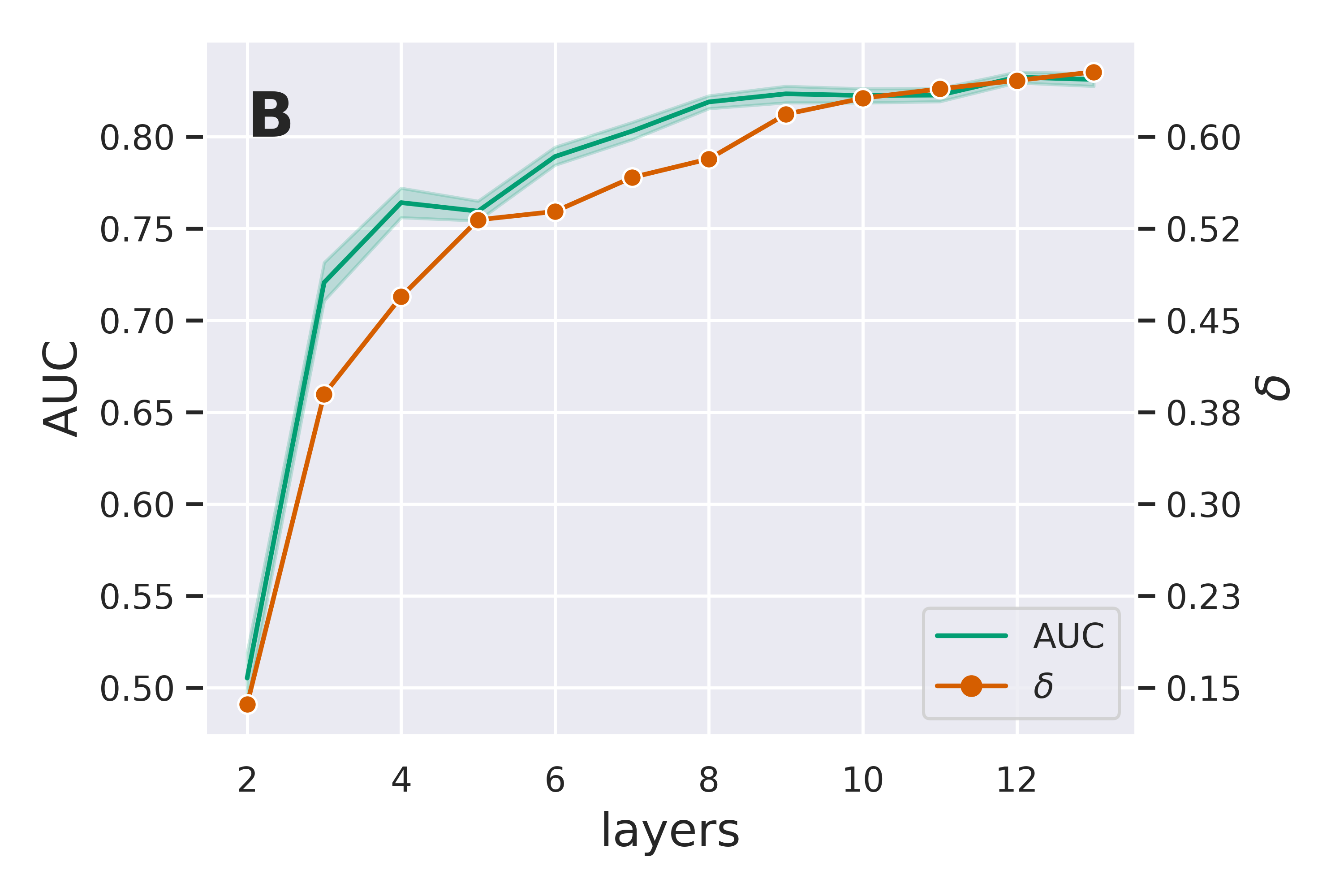}
    \caption{Variation of the AUC for inter-layer link prediction (green line) and of the parameter $\delta$ (orange line) as a function of the number of layers, on A) Drosophila and B) arXiv multiplex networks.}
    \label{fig:auc_density_comparison}
\end{figure*}

\subsection{Varying the number of layers}
As we have shown above, the accuracy of both MultiSAGE and GraphSAGE in intra-layer and inter-layer link prediction seems to depend on the number of layers of the network. To further analyze this dependence, here we carry on the following study. For the Drosophila and the arXiv datasets, we first sort the $L$ layers according to their number of nodes, so that $N_1 > N_2 > \dots > N_L$. Then, we construct a set of multiplex sub-networks, the first being formed by layers $\{1,2\}$, i.e., the two layers with the largest number of nodes, the second formed by layers $\{1,2,3\}$, and so on. Finally, the last network of the set consists in the original multiplex network, formed by all the $L$ layers. Therefore, for each network in the set, we consider the experimental setup previously described, again training and testing the algorithm over $20$ different sampling of the marked nodes. To measure the accuracy of both MultiSAGE and GraphSAGE, we adopt once again the AUC averaged over the different realizations of the embedding procedure. 

Fig.~\ref{fig:lasagna_results} shows the results obtained for the Drosophila (panel A) and the arXiv (panel B) multiplex networks. First, one can observe that, for both datasets, the accuracy of MultiSAGE in the prediction of the inter-layer links is generally higher compared to the one of GraphSAGE, for every sub-network considered. The only exception consists of the embedding of the arXiv sub-network formed by two layers, for which we find $\mathrm{AUC}_{\rm MS} = 0.51 \pm 0.03$ and $\mathrm{AUC}_{\rm GS} = 0.59 \pm 0.03$. As regards to the prediction of the intra-layer connections, GraphSAGE performs better than MultiSAGE, despite the difference between the two algorithms is not as marked as it is for the inter-layer link prediction. According to these results, our algorithm seems to display a sort of compensation effect, so that, to enhance its performance in the reconstruction of inter-layer connections, it loses in accuracy when predicting the intra-layers links. We also note that, for both datasets and for both the algorithms, the accuracy of the intra-layer link prediction does not vary much with the number of layers considered, whereas the AUC for the inter-layer link prediction clearly depends on it. Indeed, one can see that the latter increases as we add layers, up to a point where the AUC curves seem to saturate to a certain value, as it is particularly evident for the case of the Drosophila network. 

We now want to further investigate the trend of the AUC curves for the inter-layer link prediction, aiming at explaining the cause of the accuracy dependence on the number of layers. To do so, let us define the following parameter:

\begin{equation}
    \delta(L) = 1 - \frac{m_L}{\sum\limits_{l=2}^{L}(l-1)N_l} \in [0,1],
\end{equation}

where $m_L$ is the number of inter-layer links when $L$ layers are considered, and $N_l$ are the number of nodes at each layer $l$. Intuitively, this parameter corresponds to the density of inter-layer links that are not present in the network. To clarify this, let us consider $L=2$, i.e., a two-layer network. In this case, the parameter $\delta$ is simply given by 

\begin{equation}
    \delta(2) = 1 - \frac{m_2}{N_2},
\end{equation}
where $m_2$ is the number of inter-layer nodes between the two layers and $N_2$ is the number of nodes in the smallest layer, i.e., $N_2\leq N_1$. When all the nodes in layer $2$ are connected to a node on layer $1$ we have $m_2=N_2$, thus $\delta(2)=0$. However, if there are nodes on layer $2$ that do not have an inter-layer neighbor on layer $1$, one has $m_2< N_2$, which leads to $\delta(2)>0$, up to the case where no inter-layer links exist, for which $\delta(2)=1$. Note that the term $(l-1)$ within the summation comes from the fact that, when we add the $l$-th layer to the network, we could connect each of the $N_l$ nodes to $(l-1)$ nodes, one for each of the $(l-1)$ layers already present.

Fig.~\ref{fig:auc_density_comparison} shows the comparison between the AUC curves for the inter-layer link prediction and the parameter $\delta(L)$ defined above, for the Drosophila (panel A) and the arXiv (panel B) multiplex networks. As one can note, for both datasets $\delta(L)$ well correlates with the AUC curve. Such a result suggests that MultiSAGE, at least when performing a featureless network embedding, better predicts the inter-layer connectivity when the density of inter-layer links is low, i.e., when $\delta$ is closer to one. To have an intuition of the role of the link density in the algorithm accuracy, let us imagine to perform the embedding of a (almost) complete graph using GraphSAGE. As (almost) all the nodes in the network are adjacent, at each step of the embedding procedure they will be characterized by the same vector representation. As a consequence, the accuracy of the link prediction will be low, since all the connections in the test set, both positive and negative, will be predicted as positive. In the next section we will further investigate such a result, showing that the GraphSAGE (and consequently the MultiSAGE) accuracy is affected not only by the network sparsity, but also by its randomness, so that the more the graph is random the worse the performance of the algorithm is.

\subsection{The impact of the network randomness}
As we have discussed in the previous section, the accuracy of the inter-layer link prediction in the MultiSAGE algorithm seems to depend on the density of connections. To better understand such a result, here we come back to GraphSAGE, which shares the same embedding strategy as MultiSAGE, while being a simpler algorithm to examine. We hereby perform two different numerical analysis. Initially, we study how the accuracy of the link prediction changes as we randomly add links to a network, i.e., as we make the network denser. Then, we investigate the impact on GraphSAGE performance of randomly rewiring the network links, namely, we explore how making the network more random affects the link prediction accuracy.

We begin by further analyzing the role of the link density in the prediction accuracy. To carry out the study, we generate a set of networks constructed by starting from the largest layer of the arXiv dataset and adding a fraction of links $\rho$ between nonadjacent nodes. As the additional links are homogeneously distributed, we can think the newly constructed graphs as the union\footnote{The union of two graphs $G_1 = (V_1, E_1)$ and $G_2 = (V_2, E_2)$ is here defined as the graph $G = (V,E) = (V_1\cup V_2, E_1\cup E_2)$. Note that, as in the case here considered we have $V_1=V_2$, it follows that $V=V_1=V_2$.} between the original graph and an Erd\"os-R\'enyi (ER) random graph \cite{erdhos1959}, with $n_1$ nodes and connection probability $\rho$. Considering also the starting network ($\rho=0$), we account for a total of eleven networks, that we embed following the usual experimental setup, training and testing the algorithm over $20$ different sampling of the marked nodes. Again, to measure the prediction accuracy, we consider the AUC averaged over the different runs of the embedding procedure. 

\begin{figure}[t]
    \centering
    \includegraphics[width=\linewidth]{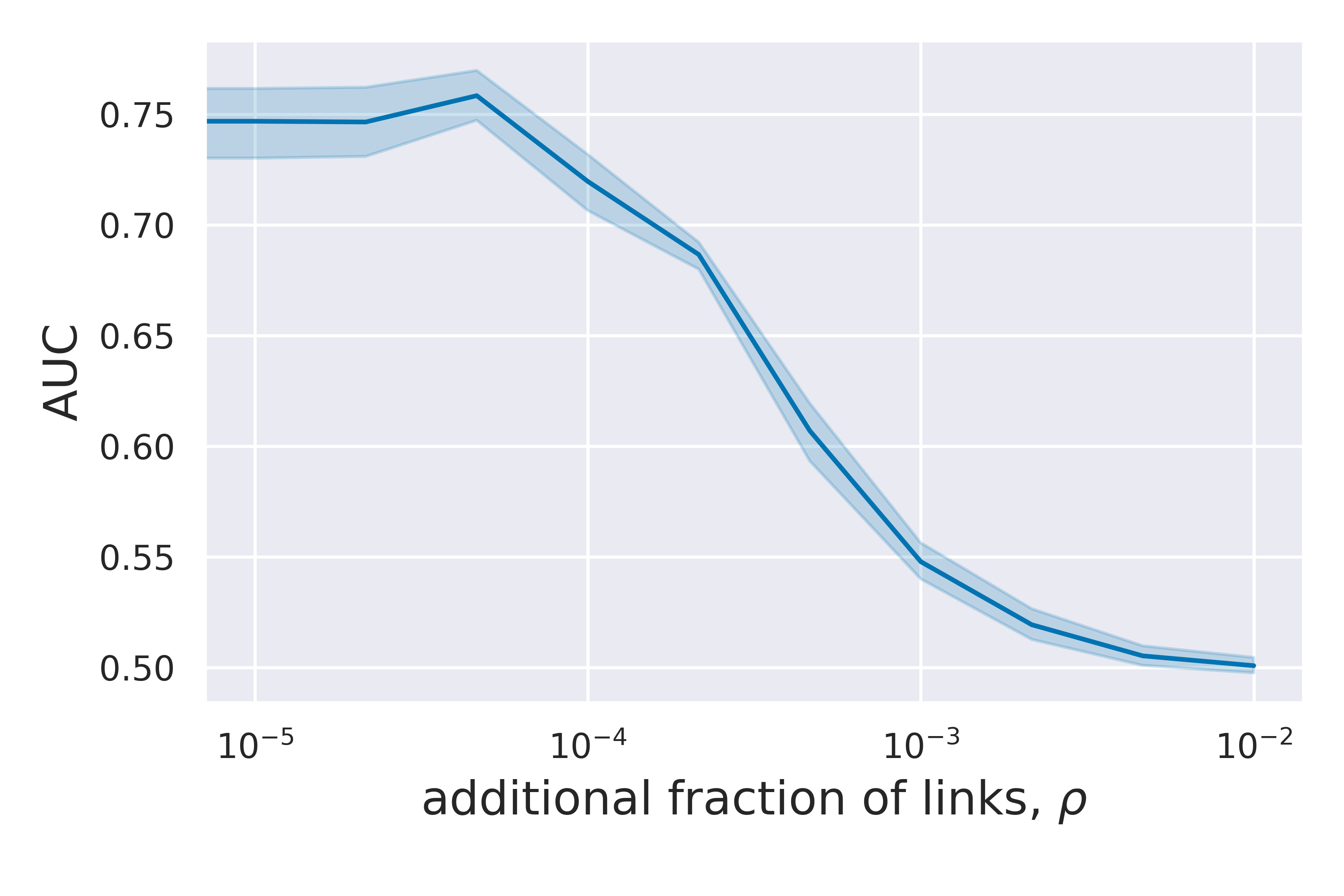}
    \caption{Variation of GraphSAGE link prediction accuracy as a function of the fraction of links $\rho$ added to the original network, consisting in the larger layer of the arXiv multiplex network. Note that the plot scale is linear-log.}
    \label{fig:auc_function_density}
\end{figure}

Fig.~\ref{fig:auc_function_density} displays how the average AUC varies as a function of the fraction of additional links $\rho$. In agreement with the results of the previous section, when the network is sparse, i.e., for $\rho<10^{-4}$, GraphSAGE is able to distinguish between positive and negative links, as shown by the high value of the AUC, which is close to $0.75$. However, when the density of links increases, the accuracy of the algorithm starts to decrease, down to $\mathrm{AUC}=0.5$ ($\rho\approx 10^{-2}$), where GraphSAGE performs as a random classifier. We can therefore conclude that, when embedding a graph with featureless nodes, GraphSAGE is able to accurately reconstruct the network connectivity only when the network itself is sparse, whereas it fails to do so for graphs that are dense. 

We now want to analyze how the network randomness influences the accuracy of the algorithm predictions. To do so, we construct a collection of networks using the Watts and Strogatz (WS) model \cite{watts1998collective}. In particular, starting from a ring lattice of $N=10^4$ nodes, where each nodes is connected to $K=4$ neighbors, two on each side, we generate the networks by rewiring the links of the lattice with a certain probability $\phi$. One crucial feature of the WS model is that we can tune the network from a regular structure ($\phi=0$) to a disordered one ($\phi=1$), while keeping the link density constant, i.e., $D=K/(N-1)$. 
Therefore, the WS model allows to study how the randomness of the network connectivity affects the accuracy of the link prediction, leaving out the contribution of edge density. Including also the regular lattice ($\phi=0$), we consider eleven networks in total. Again, we embed each network following the experimental setup described above, with $20$ different train-test splits. As an accuracy measure, we once again evaluate the AUC averaged over the different realizations of the embedding procedure.

\begin{figure}[t]
    \centering
    \includegraphics[width=\linewidth]{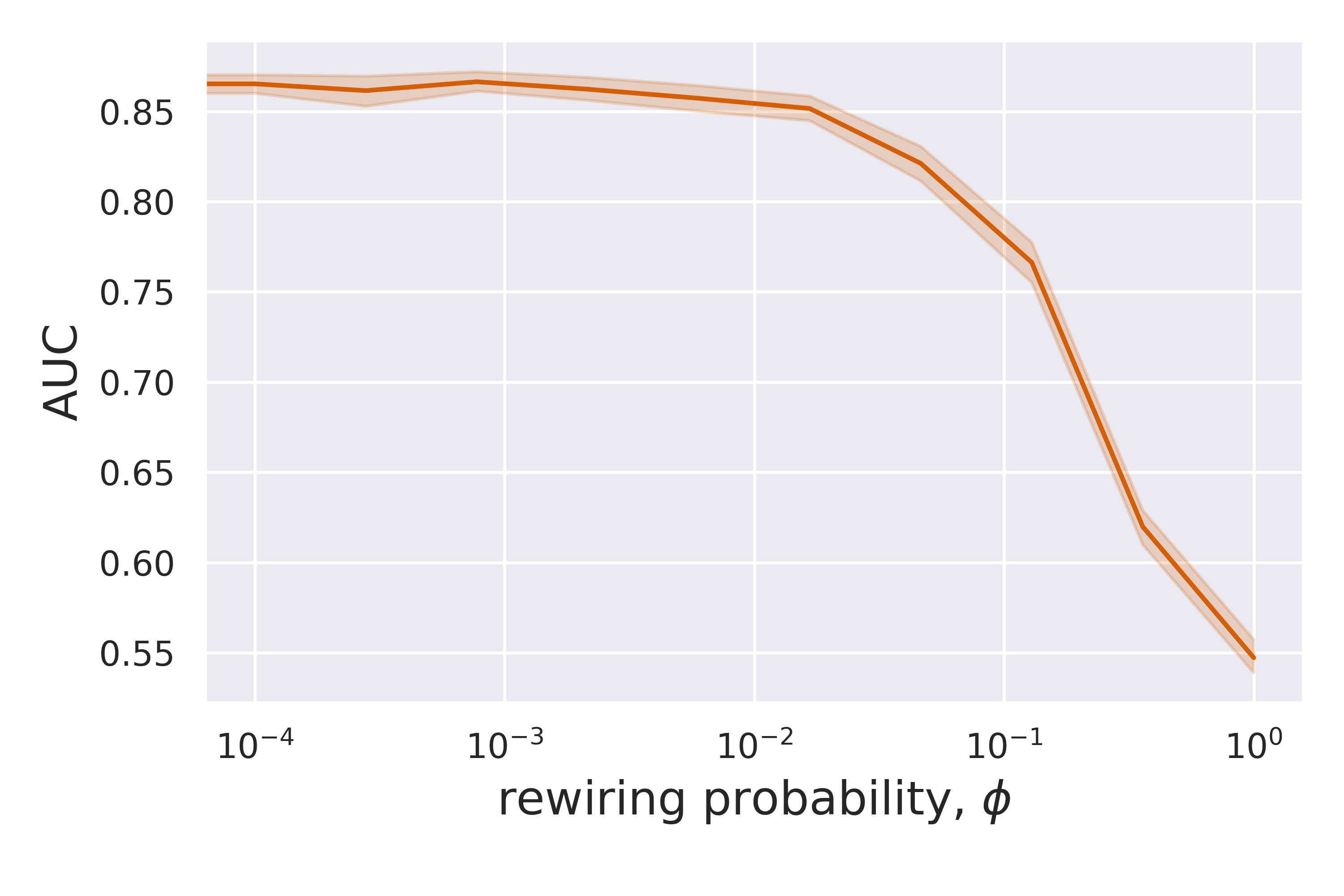}
    \caption{Variation of GraphSAGE link prediction accuracy as a function of the rewiring probability $\phi$ in the Watts and Strogatz model. Note that the plot scale is linear-log.}
    \label{fig:auc_function_probability}
\end{figure}

Fig.~\ref{fig:auc_function_probability} shows the variation of the average AUC as a function of the rewiring probability $\phi$. In particular, we observe that the accuracy of GraphSAGE in the link prediction is negatively affected by the randomness of the network. Indeed, for smaller rewiring probabilities, i.e., $\phi<10^{-2}$, the AUC is over $0.85$, meaning that the algorithm is performing well in the positive and negative links discrimination. However, for higher values of $\phi$, the accuracy of the models starts to decrease. In particular, for $\phi=1$ 
GraphSAGE almost behaves as a random classifier, as $\mathrm{AUC}=0.55$. Summing up, this analysis shows that randomness can negatively affect the performance of the algorithm. In particular, the more a network is similar to a homogeneous random graph
the worse will be the prediction accuracy. On the other hand, the results obtained suggest that networks with a regular structure or with an high clustering coefficient, such as the one generated by the WS model for small value of the rewiring probability, allows for better performances in the link prediction task. Finally, it is worth mentioning once again that such an analysis is performed in a condition where no node is provided with an external feature vector. We leave such a study for future work.

\section{Conclusion}
In this paper we have introduced an algorithm for the embedding 
of multiplex network, which distinguishes between intra- and inter-layer edges and produces reliable results, especially for the prediction of inter-layer link. We have tested the algorithm performance on different types of empirical multiplex networks. The results of our analysis clearly show that taking into account the multi-layer nature of the network positively influences the quality of the embedding. On the other hand, we also found that, either increasing the density of links, or shuffling the network edges, negatively influences the quality of the embedding.

\ifCLASSOPTIONcompsoc
  \section*{Acknowledgments}
\else
  \section*{Acknowledgment}
\fi

L.G. would like to thank C. Monti for the useful discussion and for his precious suggestions on how to perform the numerical analysis.

\ifCLASSOPTIONcaptionsoff
  \newpage
\fi

\bibliographystyle{IEEEtran}
\bibliography{biblio}

\begin{IEEEbiography}[{\includegraphics[width=1in,height=1.25in,clip,keepaspectratio]{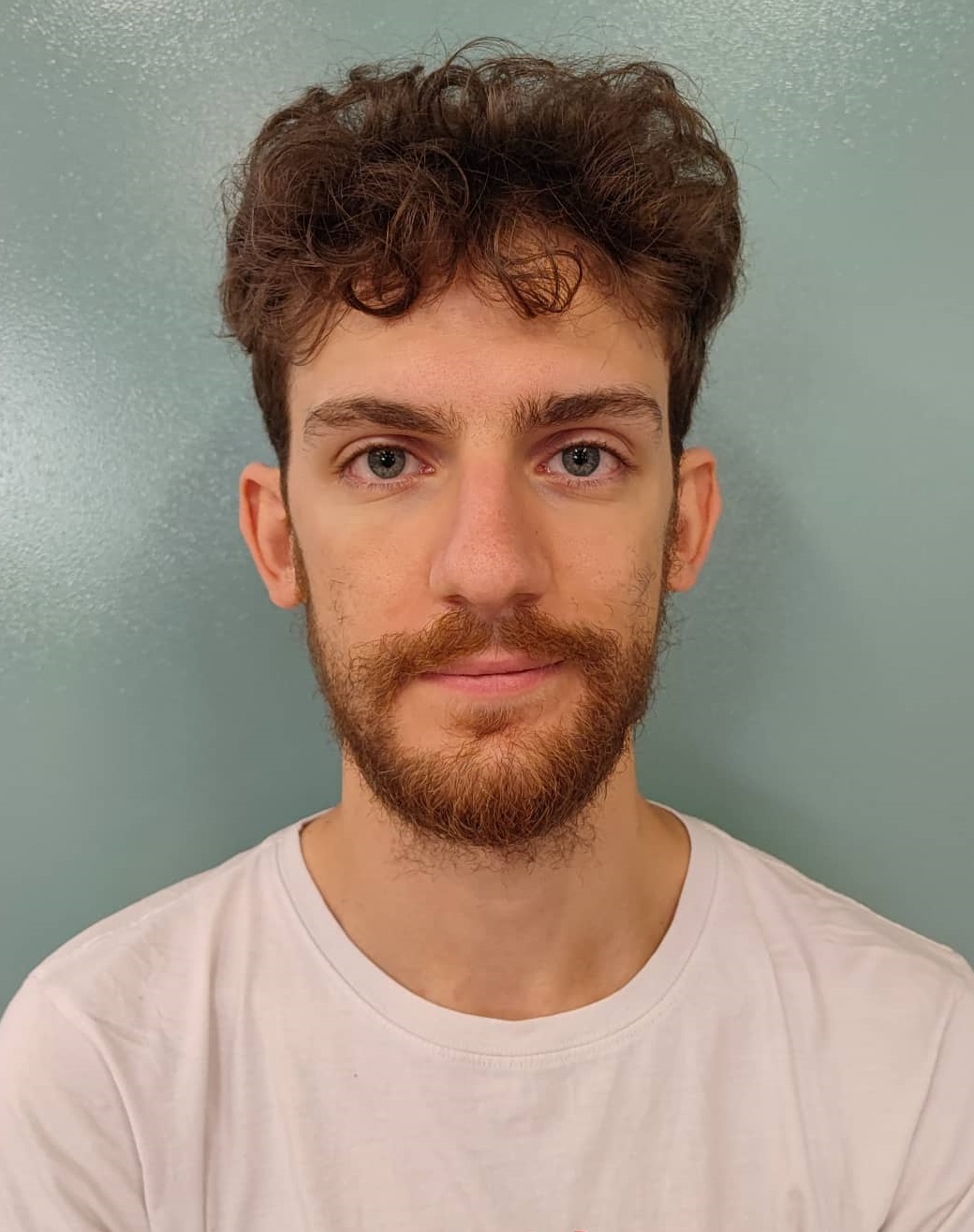}}]{Luca Gallo}
received the bachelor's degree in physics in 2016, and the master's degree in physics of complex systems in 2019, both from the University of Turin. Currently, he is a Ph.D. student in complex Systems for physical, socio-economic and life sciences at the University of Catania. His research interests include dynamical systems, complex networks, and higher-order networks.
\end{IEEEbiography}

\begin{IEEEbiography}[{\includegraphics[width=1in,height=1.25in,clip,keepaspectratio]{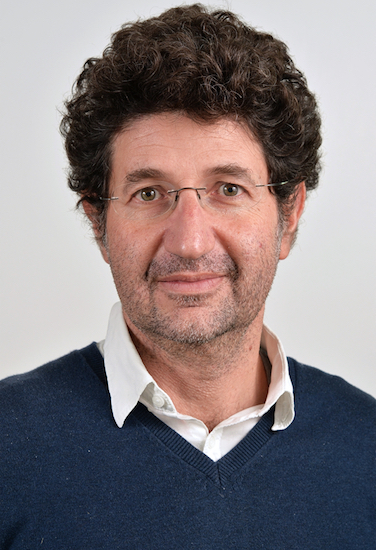}}]{Vito Latora} received the Ph.D. in Theoretical Physics in 1996 from the University of Catania, Italy. Currently, he is  
Professor of Applied Mathematics, Chair of Complex Systems, and Head of the Complex Systems and Networks Research Group in the School of Mathematical Sciences at Queen Mary University of London. He is also Professor of Physics at the University of Catania. 
His research interests include the structure and dynamics of multiplex networks, temporal networks and higher-order networks. He is interested in applications of network science to study innovation ecosystems and to understand the social and neural aspects of human creativity. He is coauthor of more than 200 scientific publications, and editor of the Journal of Complex Networks.
\end{IEEEbiography}

\begin{IEEEbiography}[{\includegraphics[width=1in,height=1.25in,clip,keepaspectratio]{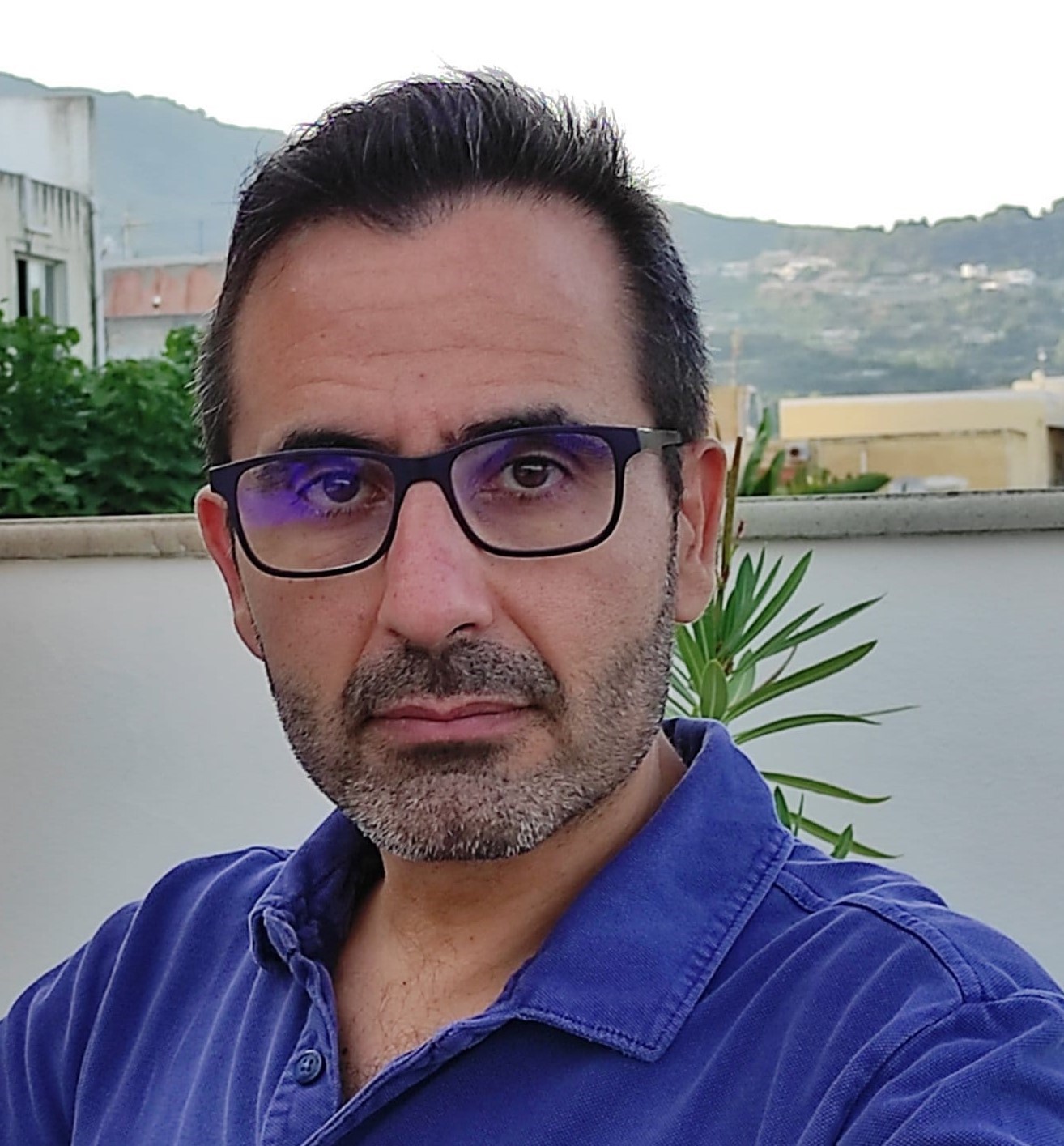}}]{Alfredo Pulvirenti} received the Ph.D. in Computer Science in 2003 from the University of Catania. Currently, he is  Associate Professor of Computer Science at the University of Catania.  His research interests mainly focuses on bioinformatics and data analysis with applications in biomedicine. In particular,  he studies methods for pathway and biological network analysis, drug repositioning, ncRNAs, design of pipelines for RNA-seq and DNA-Seq data, subgraph matching, and motif finding. He his coauthor of more than 140 scientific publications.
\end{IEEEbiography}

\end{document}